\documentclass[10pt,journal,compsoc]{IEEEtran}
\UseRawInputEncoding
\usepackage{multirow}
\usepackage{array}
\usepackage{bbding}
\usepackage{graphicx}  
\usepackage{tablefootnote}
\usepackage{flushend}

\ifCLASSOPTIONcompsoc
  \usepackage[nocompress]{cite}
\else
  \usepackage{cite}
\fi

\ifCLASSINFOpdf
\else
\fi

\begin{document}
%
\title{AI in Human-computer Gaming: Techniques, Challenges and Opportunities}
%
%
%
%

\author{Qiyue Yin,
        Jun Yang,
        Kaiqi Huang,
        Meijing Zhao,
        Wancheng Ni,\\
        Bin Liang,
        Yan Huang,
        Shu Wu,
        Liang Wang 
\IEEEcompsocitemizethanks{\IEEEcompsocthanksitem Qiyue Yin, Kaiqi Huang, Meijing Zhao, Wancheng Ni, Yan Huang, Shu Wu and Liang Wang are
with Institute of Automation, Chinese Academy of Sciences, Beijing,
China, 100190.\protect\\
E-mail: qyyin@nlpr.ia.ac.cn
\IEEEcompsocthanksitem Jun Yang and Bin Liang are with the Department of Automation, Tsinghua University, Beijing,
China, 100084.\protect\\
E-mail: yangjun603@tsinghua.edu.cn
}
}

%
%

\markboth{Journal of \LaTeX\ Class Files,~Vol.~14, No.~8, August~2015}%
{Shell \MakeLowercase{\textit{et al.}}: Bare Demo of IEEEtran.cls for Computer Society Journals}
%



\IEEEtitleabstractindextext{%
\begin{abstract}
With breakthrough of the AlphaGo, human-computer gaming AI has ushered in a big explosion, attracting more and more researchers all around the world.
As a recognized standard for testing artificial intelligence, various human-computer gaming AI systems (AIs) have been developed such as the Libratus, OpenAI Five and AlphaStar, beating professional human players.
The rapid development of human-computer gaming AIs indicate a big step of decision making intelligence, and it seems that current techniques can handle very complex human-computer games.
So, one natural question raises: what are the possible challenges of current techniques in human-computer gaming, and what are the future trends?
To answer the above question, in this paper, we survey recent successful game AIs, covering board game AIs, card game AIs, first-person shooting game AIs and real time strategy game AIs.
Through this survey, we
1) compare the main difficulties among different kinds of games and the corresponding techniques utilized for achieving professional human level AIs;
2) summarize the mainstream frameworks and techniques that can be properly relied on for developing AIs for complex human-computer gaming;
3) raise the challenges or drawbacks of current techniques in the successful AIs; and
4) try to point out future trends in human-computer gaming AIs.
Finally, we hope this brief review can provide an introduction for beginners, and inspire insights for researchers in the field of AI in human-computer gaming.
\end{abstract}

}

\maketitle

\IEEEdisplaynontitleabstractindextext

%
\IEEEpeerreviewmaketitle

\IEEEraisesectionheading{\section{Introduction}\label{sec:introduction}}

%
%
%
%
\IEEEPARstart{H}{uman}-computer gaming has a long history and has been a main stream for verifying key technologies of artificial intelligence.
Turing test \cite{Turing}, proposed in 1950, may be the first human computer gaming to judge whether the machine has human intelligence.
This inspires researchers to develop AIs that can challenge professional human players.
A typical example is a draughts AI called Chinook, which is developed in 1989 to defeat the world champion, and such a target is achieved by wining Marion Tinsley in 1994 \cite{Chinook}.
Afterwards, Deep Blue from IBM beats the chess grandmaster Garry Kasparov in 1997, making a new era in the history of human-computer gaming \cite{deepBlue}.

Recent years, we witness the rapid development of human-computer gaming AIs, from the DQN agent \cite{Atari}, AlphaGo\cite{AlphaGo}, Libratus\cite{Libratus}, OpenAI Five\cite{OpenAIFive} to AlphaStar\cite{AlphaStar}.
Those AIs defeat professional human players in certain games by a combination of modern techniques, indicating a big step of the decision making intelligence \cite{Agent57,HideAndSeek,Pluribus}.
For example, AlphaGo Zero\cite{AlphaGoZero}, utilizing Monte Carlo tree search, self-play and deep learning, defeats dozens of professional go players, representing powerful techniques for lager state perfect information games.
OpenAI Five \cite{OpenAIFive}, using self-play, deep reinforcement learning and continual transfer via surgery, becomes the first AI to beat the world champions at an esports game, displaying useful techniques for complex imperfect information games.

After success of the AlphaStar and the OpenAI Five, which reach professional human player level in the games StarCraft and Dota2, respectively, it seems that current techniques can solve very complex games.
Specially, the breakthrough of the most recent human-computer gaming AIs for games such as the honor of kings \cite{Honor}, Mahjong \cite{Suphx} obey similar frameworks of AlphaStar and OpenAI Five, indicating a certain degree of universality of current techniques.
So, one natural question raises: what are the possible challenges of current techniques in human-computer gaming, and what are the future trends?
This paper aims to review recent successful human-computer gaming AIs, and try to answer the question through a thorough analysis of current techniques.

\begin{figure*}
   \begin{center}
   \includegraphics[width=0.9\textwidth]{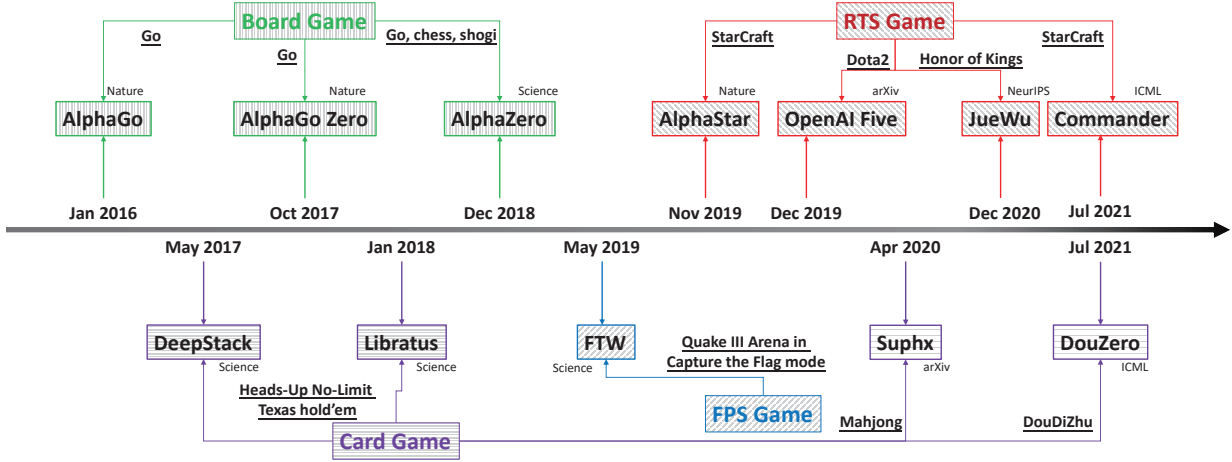}
   \end{center}
   \caption{Games and AIs surveyed in this paper.}
   \label{AIs}
\end{figure*}

Based on current breakthrough of human-computer gaming AIs (most published in journals such as Science and Nature), we survey four typical types of games, i.e., board games with Go; card games such as heads-up no-limit Texas hold¡¯em (HUNL), DouDiZhu and Mahjong; first person shooting games (FPS) with Quake III Arena in Capture the Flag (CTF); real time strategy games (RTS) with StarCraft, Dota2 and Honor of Kings.
The corresponding AIs cover AlphaGo\cite{AlphaGo}, AlphaGo Zero\cite{AlphaGoZero}, AlphaZero\cite{AlphaZero}, Libratus\cite{Libratus}, DeepStack\cite{DeepStack}, DouZero\cite{DouZero}, Suphx\cite{Suphx}, FTW\cite{CTF}, AlphaStar\cite{AlphaStar}, OpenAI Five\cite{OpenAIFive}, JueWu\footnote{A name known by the public.}\cite{Honor} and Commander\cite{Commander}.
A brief summary is displayed in figure \ref{AIs}.

\begin{table*}
\begin{center}
\caption{Characteristics of four typical kinds of games.}
\label{gamesIntro}
\scalebox{1}{
\begin{tabular}{c c c c c c c c c}
\hline
\multirow{2}{*}{Games}       & \multicolumn{1}{c}{Boad Games}    & \multicolumn{3}{c}{Card Games}    & \multicolumn{1}{c}{FPS Games}     & \multicolumn{3}{c}{RTS Games}    \\
                        & Go series   & HUNL & DouDiZhu  & Mahjong   & CTF   & StarCraft & Dota2 & Honor of Kings    \\
\hline
Imperfect information   & \XSolid     & \Checkmark     & \Checkmark     & \Checkmark     & \Checkmark     & \Checkmark     & \Checkmark     & \Checkmark  \\
Long time horizon       & \Checkmark  & \XSolid     & \XSolid     & \XSolid     & \Checkmark     & \Checkmark     & \Checkmark     & \Checkmark  \\
In-transitive game      & \Checkmark     & \Checkmark     & \Checkmark   & \Checkmark   & \Checkmark   & \Checkmark   & \Checkmark   & \Checkmark     \\
Multi-agent cooperation & \XSolid     & \XSolid     & \Checkmark    & \XSolid  & \Checkmark     & \Checkmark    & \Checkmark    & \Checkmark  \\
\hline
\end{tabular}}
\end{center}
\end{table*}

The rest of the paper is organized as follows. In Section 2, we describe games and AIs covered in this paper. Sections 3-6 elaborate the AIs for board games, card games, FPS games and RTS games, respectively. In Section 7, we summarize and compare different techniques utilized. In Section 8, we show the challenges in current game AIs, which maybe the future research direction of this field. Finally, we conclude the paper in Section 9.

\section{Typical Games and AIs}
Based on recent progresses of human-computer gaming AIs, this paper reviews four types of games and their corresponding AIs, i.e., board games, card games, FPS games and RTS games. To measure how hard a game is to develop professional human level AI, we extract several key factors that challenge the intelligent decision making \cite{Yin}, which are displayed in Table \ref{gamesIntro}.

\textbf{Imperfect information}. Except for the board games, almost all the card games, FPS games and RTS games are imperfect information games, which means players do not know exactly how they come to the current states, e.g, current face in HUNL.
Accordingly, players need to make decisions under partial observations.
This leads to more than one nodes in an information set if the game is expanded into a tree.
For example, the average information sets for card games HUNL and Mahjong are $10^3$ and $10^{15}$, respectively.
Compared with perfect information games such as go, a subgame in an imperfect information game cannot be solved isolated from each other \cite{Imperfect}, which makes solving Nash equilibrium of imperfect information games more difficult\cite{Hard}.

\textbf{Long time horizon}. In real time games, such as StarCraft, Dota2 and Honot of Kings, a game lasts several minutes and even more than an hour.
Accordingly, an AI needs to make thousands of decisions.
For example, Dota 2 games run at 30 fps for about 45 minutes, which results to approximately 20,000 steps in a game if making a decision every four frames.
In contrast, players in the board games and card games usually make much less decisions.
The long time horizon leads to an exponential increase in the number of decision points, which brings in a series of problems such as exploration and exploitation when optimizing a strategy.

\textbf{In-transitive game}. If performance of different players are transitive, a game is called a transitive game \cite{Transitive}.
Mathematically, if $v_t$ can beat $v_{t-1}$ and $v_{t+1}$ can beat $v_{t}$, $v_{t+1}$ outperforms $v_{t-1}$. Then a game is strictly transitive.
However, most games in real world are in-transitive.
For example, in a simple game "Rock-Paper-Scissor", the strategy is in-transitive or cyclic.
Specially, it is common that most games consist of transitive and in-transitive parts, i.e., obey the spining tops structure \cite{gameStructure}.
The in-transitive characteristic makes standardized self-play technique, widely used for agent ability evolution, fail to iteratively reach the Nash equilibrium strategy.

\begin{figure*}
   \begin{center}
   \includegraphics[width=0.875\textwidth]{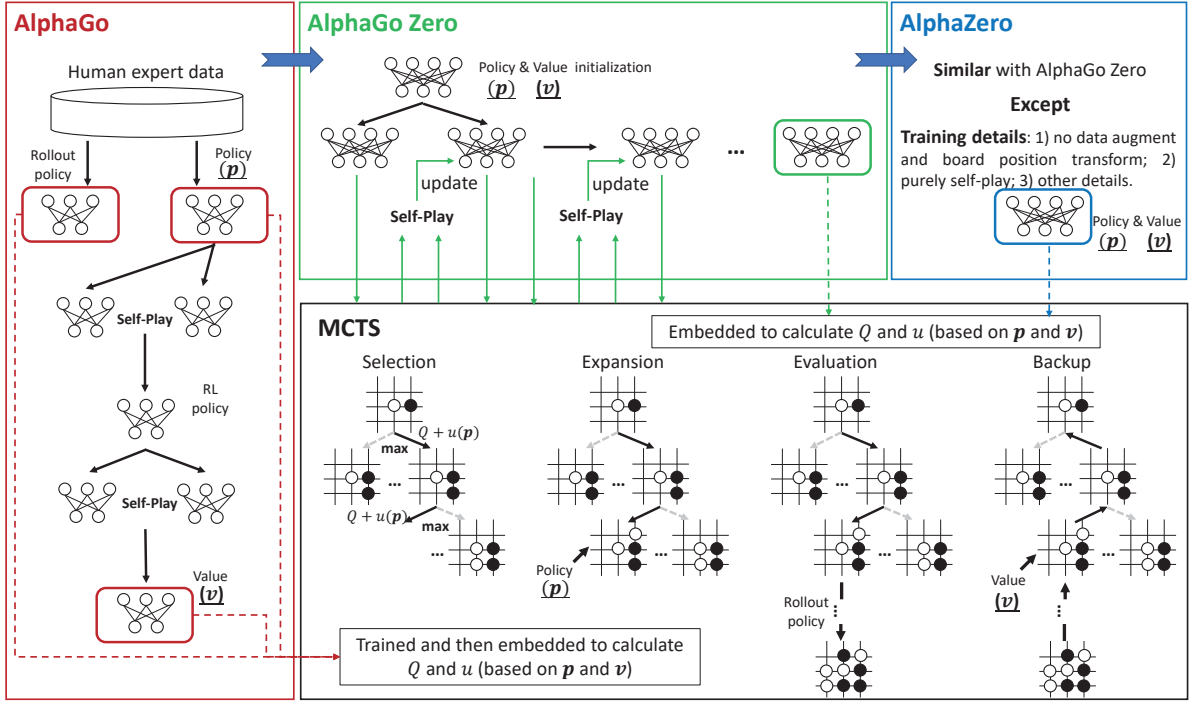}
   \end{center}
   \caption{A brief framework of AlphaGo series.}
   \label{Go}
\end{figure*}

\textbf{Multi-agent cooperation}. Most board games and card games are purely competitive, where no cooperation between players is required.
An exception is DouDizhu, which needs two Peasants players playing as a team to fight against the Landlord player.
In contrast, almost all the real time games, i.e., FPS games and RTS games, rely on players' cooperation to win the game.
For example, Five players in Dota2 and Honor of Kings form a camp to fight against another camp.
Even though StarCraft is a two-payer competitive game, each player needs to control a large number of units, which need to be well cooperated for a win.
Overall, how to obtain the Nash equilibrium strategy or a better learned strategy under the multi-agent cooperation is a hard problem, because specially designed agent interaction or alignment needs to be considered.

In summary, different games share different characteristics and aim to find different kinds of solutions, so distinct learning strategies are developed to build AI systems. In this paper, the AIs cover: AlphaGo, AlphaGo Zero, AlphaZero for board game Go; Libratus, DeepStack, DouZero and Suphx for card games HUNL, DouDiZhu and Mahjong, respectively; FTW for FPS game Quake III Arena in Capture the Flag model; AlphaStar, Commander, OpenAI Five and JueWu for StarCraft, Dota2 and Honor of Kings, respectively.

\section{Board Game AIs}\label{sec3}
AlphaGo series consist of AlphaGo, AlphaGo Zero and AlphaZero.
AlphaGo, come out in 2015, beats European Go champion Fan Hui by 5:0, which is the first time that AI wins professional players in full size game Go without Renzi.
Afterwards, an advanced version called AlphaGo Zero is developed by using different training frameworks, which needs no prior professional human confrontation data and reaches superhuman performance.
AlphaZero, using similar training framework with AlphaGo Zero, and serving as an exploration of general reinforcement learning algorithm, that masters Go along with another two board games chess and shogi.
A brief summarization is shown in figure \ref{Go}.

\subsection{MCTS for AlphaGo Series}
One of the key factors of AlphaGo series is MCTS, which is a typical tree search based method \cite{MCTS,MCTSD}.
Generally, a simulation of MCTS consists of four steps, which is repeated hundreds and thousands of times for one step decision.
The four steps consist of selection, expansion, evaluation and backup, which are operated in a tree as shown in lower right corner of Figure \ref{Go}.
Selection selects one leaf node starting from the root node, i.e., the state where an action needs to be decided, based on the evaluation of the nodes in the tree.
Expansion expands the tree by adding a new node.
Starting from the expanded node, a rollout is performed to obtain a value for the node, which is used to update the values of all nodes in the tree.

In the AlphaGo series, traditional MCTS is improved via deep learning to limit the width and depth of the search, so as to handle the huge game tree complexity.
Firstly, in the selection stage, a node is selected based on the sum of action value $Q$ and a bonus $u(p)$.
The action value is the average node values of all simulations, and the node value is evaluation of a node based on predication of value network and rollout results based on rollout network.
The bonus is proportional to the policy value (probability of selecting points in Go) calculated via the policy network, but inversely proportional to the visit count.
Secondly, in the expansion stage, a node is expanded and its value is initialized through the policy value.
Finally, when making an estimate of the expanded node, rollout results based on rollout network and predicted results based on value network are combined.
Noted in AlphaGo Zero and AlphaZero, rollout is removed, and the evaluation of expanded node is based solely on prediction results of value network.

\subsection{Training Differences for AlphaGo Series}
\subsubsection{Training framework of AlphaGo}
Training of AlphaGo consists of several steps. Firstly, a supervised learning (SL) policy network and a rollout policy network are trained with human expert data, which outputs the probability of next move position based on 160,000 games played by KGS 6 to 9 dan human players.
The differences between the SL policy and rollout policy are the neural network architecture and features.
With the above high quality data, a very good initiation of the SL policy network is obtained, which reaches Amateur level, i.e., about Amateur 3 dan (d).

With the trained SL policy network, a reinforcement learning (RL) policy network is initialized and then improved through self-play, which uses network of the current version to fight against its previous versions.
Based on conventional policy gradient method to maximize the wining signal, RL policy network reaches better performance than SL network, i.e., RL policy obtains 80\% winning rate against SL policy.

In the third step of AlphaGo, a value network is trained to evaluate state.
Specially, a dataset consists of 30 million state-outcome pairs is collocated through self-play of RL network.
Then, a regression task is developed by minimizing the mean squared error between the predicted result of value network and the corresponding outcome (win or loss signal).
With the value network, MCTS can reach a better performance than just using SL network.
Finally, well trained SL policy, value network and rollout network are embedded into MCTS, which reaches professional level of 1 to 3 dan (p).

\subsubsection{Training framework of AlphaGo Zero and AlphaZero}
Unlike AlphaGo, whose policy network and value network are trained through supervised learning and self-play between the policy networks, AlphaGo Zero trains policy and value networks through self-play of MCST embedded in the current version of the networks.
AlphaZero shares the same training framework with AlphaGo Zero.
Overall, they consist of two alternating repetition steps: automatically generating data; policy and value networks training.

When generating training data, self-play of MCTS is performed.
MCTS embedded in the current policy and value networks is used to select each move for the two players at each state.
Generally, MCTS selects an action based on the maximum count, but AlphaGo Zero makes it a probability to explore more actions through normalizing the count.
Accordingly, state-move probability pairs are stored.
Finally, when a game ends, the wining signal (+1 or -1) is recorded for value network training.

Relying on above collected state-move probability and wining signal, the policy and value networks are trained.
More specifically, the distance between predicted probability of policy network and collected probability for each state is minimized.
Besides, the distance between predicted value of value network and the winning signal is minimized.
The overall optimizing objective also contains an $L2$ weight regularization to prevent overfitting.

\subsubsection{Training differences}
Based on MCTS, deep learning, reinforcement learning and self-play are nicely evolved in AlphaGo series, as shown in figure \ref{Go}.
The main difference is training frameworks utilized, which is elaborated in the following.
To sum up, AlphaGo utilizes human expert data to initialize policy network, based on which, self-play between policy networks is performed to train the value network, and the trained networks are embedded into MCTS for decision making.
However, AlphaGo Zero uses no human expert data, and trains the policy and value networks based on data generated through self-play of MCTS embedded in current version of policy and value networks.
AlphaZero shares the same training framework with AlphaGo Zero, except several small training settings.

Apart form the training framework, there are several factors AlphaGo Zero differs from AlphaGo.
Firstly, no rollout policy network is used to evaluate the expanded node, and no human expert data are utilized for deep neural networks training.
Secondly, policy and value networks in AlphaGo Zero share most parameters (convolutional layers) instead of two separate networks, which shows better Elo rating.
What's more, residual blocks, as a powerful modular, is utilized in AlphaGo Zero, and it shows much better performance than just using convolutional
blocks as in AlphaGo.
Finally, the input of AlphaGo Zero is $19\times19\times17$ image stack, which rarely uses human engineering features compared with AlphaGo, such as the designed ladder capture and ladder escape features.

AlphaZero aims to develop a more general reinforcement learning algorithm for various board games such as Go, chess and Shogi.
Since rules of chess and Shogi are very different from Go, AlphaZero makes several changes of training details to fit the above goal.
As for the game Go, there are two main training details that are different with AlphaGo Zero.
Firstly, no data augment and transformation such as rotation or reflection of the positions are applied.
Secondly, AlphaZero uses a pure self-training framework by maintaining only a single neural network instead of saving a better model in each iteration of training.

\section{Card Game AIs}\label{sec4}

\begin{figure*}
   \begin{center}
   \includegraphics[width=0.9\textwidth]{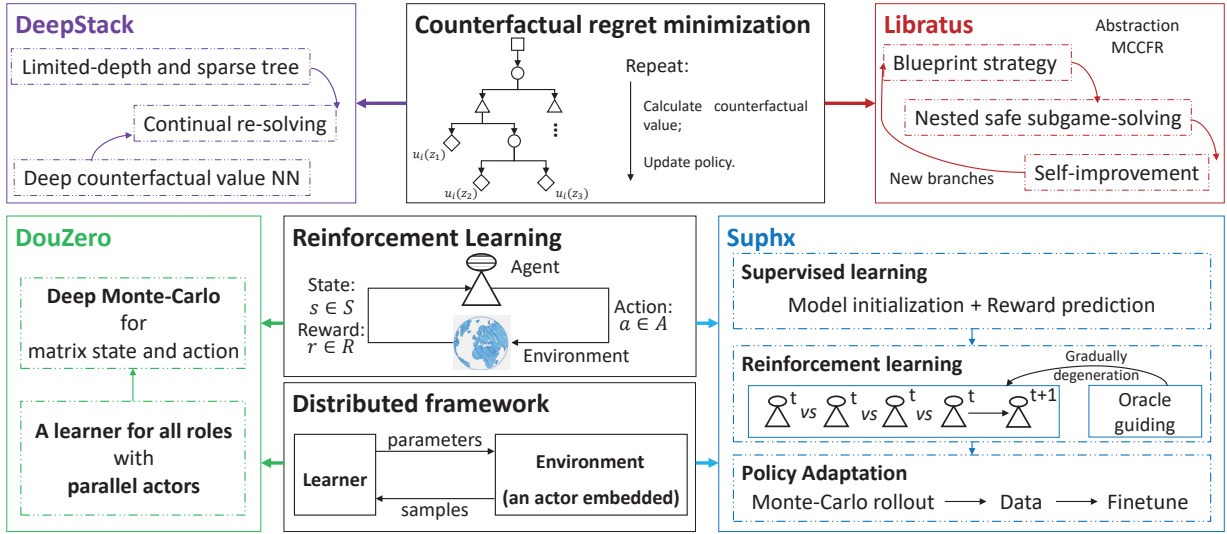}
   \end{center}
   \caption{A brief framework of Card game AIs.}
   \label{DouZero}
\end{figure*}

Card game, as a typical in-perfect information game, has been a long-standing challenge for artificial intelligence.
DeepStack and Libratus are two typical AI systems that defeat professional poker players in HUNL.
They share basic techniques, i.e, CFR, which are both theoretically sound.
Afterwards, researcher are focusing Mahjong and DouDiZhu, which raise new challenges for artificial intelligence.
Suphx, developed by Microsoft Research Asia, is the first AI system that outperforms most top human players in Mahjong.
DouZero, designed for DouDiZhu, is an AI system that was ranked the first in the Botzone leaderboard among 344 AI agents.
A brief introduction is shown in figure \ref{DouZero}.

\subsection{DeepStack and Libratus for HUNL}
HUNL is one of most popular poker games all around the world, and plenty of world-level competitions are hold every year such as the World Series of Poker (WSOP).
Before the DeepStack and Libratus came out, HUNL is a primary benchmark and challenge of imperfect information game with no AIs have defeated professional players.
\subsubsection{CFR for DeepStack and Libratus}
Since proposed in 2007, CFR has been introduced in poker games.
CFR minimizes counterfactual regret for large extensive games, which can be used to compute a Nash equilibrium.
Generally, it decomposes regret of an extensive game into a set of additive regret terms on information sets that can be minimized independently.
Due to large cost of time and space, basic CFR is not applicable for HUNL, which is far more complex than limited poker.
Various improved CFR approaches have been developed considering improving computing speed or compressing the required storage space \cite{MCCFR,CFRD}.
For example, based on  CFR, continue-resolving, and safe and nested subgame solving, are key factors for success of the DeepStack and Libratus, respectively.

\subsubsection{Training for DeepStack}
Key of training for DeepStack is continual re-solving, which is assisted by depth-limited look-ahead via deep learning and sparse look-ahead trees.
Re-solving, begins with a strategy, and reconstructs the strategy by resolving every time an decision is required.
To accomplish this at any decision point, DeepStack maintains a player's own range and opponent counterfactual values.
Giving three specific updating rules on own action, chance action and opponent action, it ensures that opponent counterfactual values are properly bounded.
A very important characteristic is no requirements for knowledge of opponent action and range to update above values, which makes DeepStack very efficient.

However, purely re-resolving is intractable because of the deep depth of game tree in HUNL.
To handle this problem, Deepstack restricts the depth of the subtree via intuition.
A counterfactual value function is trained with deep neural networks and utilized for estimating how valuable holding certain cards.
Moreover, by limiting actions to be fold, call, two or three bet actions and all-in, the resolved games are reduced to have about $10^7$ decision points, largely reduced compared to $10^{160}$ decision points for the whole game.
Based on such abstraction, DeepStack can make a decision with no more than 5 second under a machine with a single NVIDIA GeForce GTX 1080 graphics card.

\subsubsection{Training for Libratus}
Training of Libratus needs no expert domain knowledge and consists of three main steps: building a blueprint strategy, nested safe subgame solving and self-improvement.
Blueprint strategy is solved by an improved version of CRF, i.e, Monte Carlo CFR (MCCRF), for an abstracted game, which provides a strategy for early rounds of the game and an approximation for latter rounds.
As for the abstraction, certain bet sizes are abstracted based on an application-independent parameter-optimization algorithm.
However, not card abstraction on the first and second betting rounds are adopted, where decision strategy is purely based on blueprint strategy.

Nested safe subgame solving is used in the third and fourth betting rounds, which provides a real time solution for a more detailed abstraction of the game tree.
The abstraction in the blueprint is relaxed instead of rounding the bet size to the nearest size.
Libratus will make a distinct strategy in response to off-tree actions.
Nested safe subgame solving ensures that new strategy for the subgame improve blueprint strategy by making the opponent worse off no matter what cards she is holding.
Finally, Self-improvement computes a game-theoretic strategy for branches that are added based on actual moves of opponents.

\subsubsection{Training differences}
Intuitively, DeepStack solves the subtree based on re-solving assisted by deep neural networks for counterfactual values prediction, whereas, Libratus utilizes a nested safe subgame solving strategy to improve the original abstraction based strategy.
Both methods use estimated value instead of the upper bounds value of the opponent, but libratus claims that DeepStack does not share its improvement of de-emphasizing hands.

Libratus plays the first two rounds based on precomputed blueprint strategy, which makes big abstraction of opponent actions.
However, DeepStack re-solves each subgame no matter what rounds it is now deciding, making it more flexible of dealing with opponent off-tree actions.
To make Libratus more powerful handling off-tree opponent bet sizes in the first two rounds, a self-play improvement modular is designed based on actual moves of opponent, which can largely remedy defects.

\subsection{Suphx and DouZero for Mahjong and DouDiZhu}
Unlike HUNL, Mahjong has different types of actions and the regular order of plays can be interrupted, making the game tree consisting of huge number of paths between the consecutive actions of a player.
This leads the successful MCTS and CFR based techniques for Go and HUNL not a best choice.
Similarly, unlike HUNL, the actions of DouDiZhu is complex and can not be abstracted, making tree search based techniques such as MCTS and CFR hard to be applied.
To this end, Suphx and DouZero adopt deep reinforcement learning as basic tools for AI development, which aims to reach high-level perfomance and cares little about characteristics of the solution such as the Nash equilibrium.

\subsubsection{Basic techniques for Suphx and DouZero}
Reinforcement learning (RL) is a typical type of machine learning, which becomes one of most important decision-making techniques since the breakthrough of AlphaGo \cite{RLSurvey1}.
Generally, RL follows the framework of policy evaluation and policy improvement by interacting with the environment.
Because of the trial and error mechanism, RL requires a huge amount of data for policy learning, leading to sample efficient problem \cite{Sample,ModelRL}.
Distributed training, utilizes multiple machines for learning a task, is now combined with RL for alleviating the above problem \cite{Disributed,Disributed1}.

Nair et al \cite{DisDQN} proposed the first massively distributed architecture for RL, which consists of four components.
The first part is parallel actors, which are used to interact with environment and generate data;
The second component is parallel learners that consume data for policy training;
The third and fourth parts are distributed neural network and store of experience to connect the actor and learner.
Based on the above framework, a number of advanced distributed reinforcement learning frameworks are developed, and data throughput is largely improved \cite{IMPALA,SeedRL,ACME}.
In Suphx and DouZero, distributed learning is adopted to accelerate RL training, where multiple rollouts are paralleled performed to collect data.

\subsubsection{Training for Suphx}
Suphx is a hybrid learning system, which consists of rule-based wining model and five learning-based models to form the decision flow.
Generally, training of the five learning-based models contains three major steps: supervised learning, self-play reinforcement learning and a run-time policy adaptation.

Supervised learning is performed utilizing state-action pairs collected from human players in Tenhou platform, and then act as initialization for the self-play reinforcement learning stage.
Usually, each game consists of multiple rounds and the final reward signal is obtained by accumulating all the round scores, so it is hard to guild reinforcement learning in each round because some players may tactically lose several rounds to win the game.
In Suphx, such problem is solved by using a GRU network to predict feedbacks of each round.
More specifically, data of top human players are collected as reward and a regression based objective is constructed between past and present round information and the final game reward.
When performing reinforcement learning, such predication is served as the intermediate reward for each round in a game.

In reinforcement learning stage, considering learning is slow facing the rich hidden information in Mahjong, Suphx proposed a method called oracle guiding.
Firstly, an oracle agent is trained by using all the perfect information, i.e., private tiles of all the players and the tiles in the wall.
Since a simple knowledge distillation method does not work because it is hard for a normal agent with very limited information to mimic the oracle agent,
Suphx gradually drops out the perfect features so that the oracle agent can slowly degenerate to the normal agent.

Run-time policy adaptation is utilized so that the learned policy can be properly adapted based on the tiles of current round.
The motivation comes from human player, who will act very different based on different tiles in the beginning of each round.
A parametric Monte-Carlo policy adaption approach is proposed, which consist of two steps.
Firstly, Suphx simulates multiple games by self-play using previously trained policy at the beginning of a round, with which trajectories are collected.
Then gradient updates are performed using the about data for policy finetune.
Based on the experimental results, the simulation does not to be very large, and in every round, the policy adaption can be adopted.

\subsubsection{Training for DouZero}
In DouZero, a deep Monte-Carlo method is developed with specially designed matrix-form state and action spaces.
Since there are up to 27,472 possible actions for a player, a matrix-form action representation provides a beautiful way to encode and more importantly reason about unseen actions.
This is one of key factors that DouZero can handle huge action space.
Considering MC approaches are usually inefficient because of its high variance issue, DouZero utilizes distributed training to parallelize the data generation part.
Specially, a lot of actors are raised with each maintains local networks of the three players and generates episode trajectories, based on which, a learner of global networks for the three players are trained.
Overall, training algorithm of DouZero is simple and efficient, and the authors show classic MC methods can be properly designed to deal with games with a complex action space.

\subsubsection{Training differences}
Apart from utilizing reinforcement learning algorithms and distributed framework for training acceleration, training frameworks of the DouZero and Suphx are very different.
Firstly, training of Suphx is a complex and multi-stage system, whereas training of DouZero is relatively simple with a distributed deep MC method.
In Suphx, data from top human player are required for network initialization and round reward predication, based on which, it outperforms most top human players in Mahjong.
However, in DouZero, no human data is required, and networks for different players are trained from scratch, based on which, it ranks the first in the Botzone leaderboard among 344 AI programs.

\section{First-Person Shooting Game AIs}\label{sec5}
\begin{figure*}
   \begin{center}
   \includegraphics[width=0.9\textwidth]{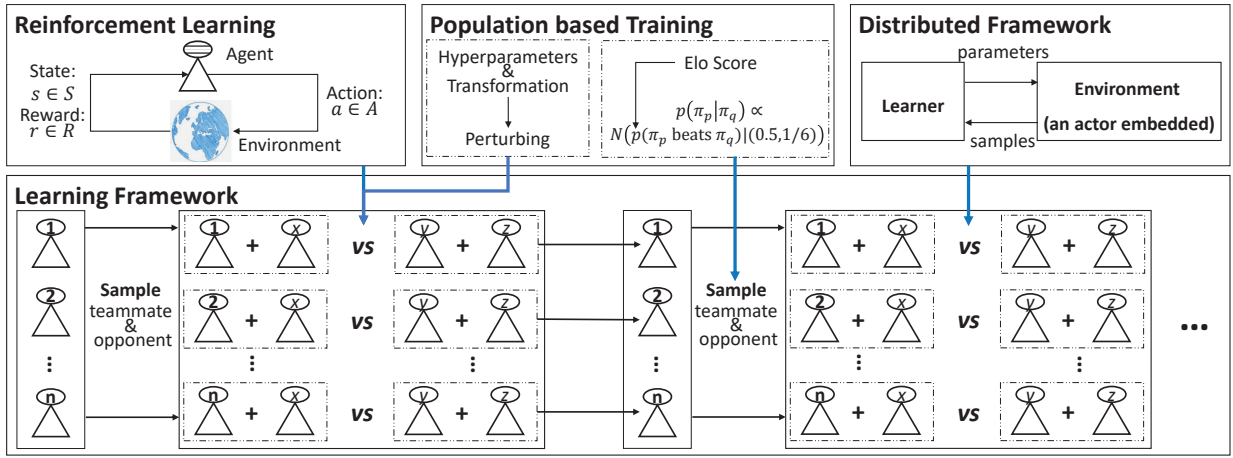}
   \end{center}
   \caption{A brief framework of FTW for game CTF.}
   \label{CTF}
\end{figure*}

Quake III Arena in Capture the Flag (CTF) mode is a typical three-dimensional multiplayer first-person video game, where two opposing teams are fighting against each other in in-door or out-door maps.
As we will see in the next section, settings for CTF are very different from current multi-player video games.
More specifically, agents in CTF cannot access the state of other players, and agents in a team cannot communicate with each other, making such an environment a very good testbed for learning agents to emerge communication and adapt to zero-shot generation.
Zero-shot means an agent cooperated or confronted is not the agent trained, which can be human players and arbitrary AI agents.
Based sorely on pixels and game points like human as input, the learned agent FTW reaches the strong human-level performance.
A brief introduction is shown in figure \ref{CTF}.

\subsection{Learning Framework}
The aim of FTW is to train agents that can adapt to the variability of maps, number of agents, and choice of teammates and opponents.
To achieve such high scalability, conventional self-play methods are claimed to be unstable, and those approaches in their basic form cannot support concurrent training, which are important for scalability.
To handle the problems, FTW trains in parallel a population of agents, where each agent is trained based on distributed reinforcement learning with experiences collected by dynamically selected teammates and opponents.
Moreover, an online evolutionary algorithm is developed guiding agents learning, so as to directing the population.
The above processes are called population based training, which will be elaborated in the following subsection.

Considering the global reward is sparse for FTW, which lasts for 4500 frames.
FTW learns several intermediate rewards to accelerate training.
A key problem of learning such rewards is to ensure the optimization of intermediate rewards promotes the policy optimization for chasing global rewards.
Such problem is solved by a specially designed joint maximization objective, where inner optimization optimizes intermediate rewards through distributed reinforcement learning, and outer optimization, regarded as a meta-game, is optimized through population based training for transformation between intermediate reward and global reward.

Another specific aspect of FTW lies in its neural network design.
Due to partial observation of the agent, FTW follows the idea of reinforcement learning as probabilistic inference.
Accordingly, a hierarchical LSTM network with different timescales is developed, where the LSTM with fast timescale generates hidden states and enhanced by the LSTM with slow time scale.
Hidden states of LSTM with fast timescale then severs as the variational posterior for the final action selection.

\begin{figure*}
   \begin{center}
   \includegraphics[width=0.9\textwidth]{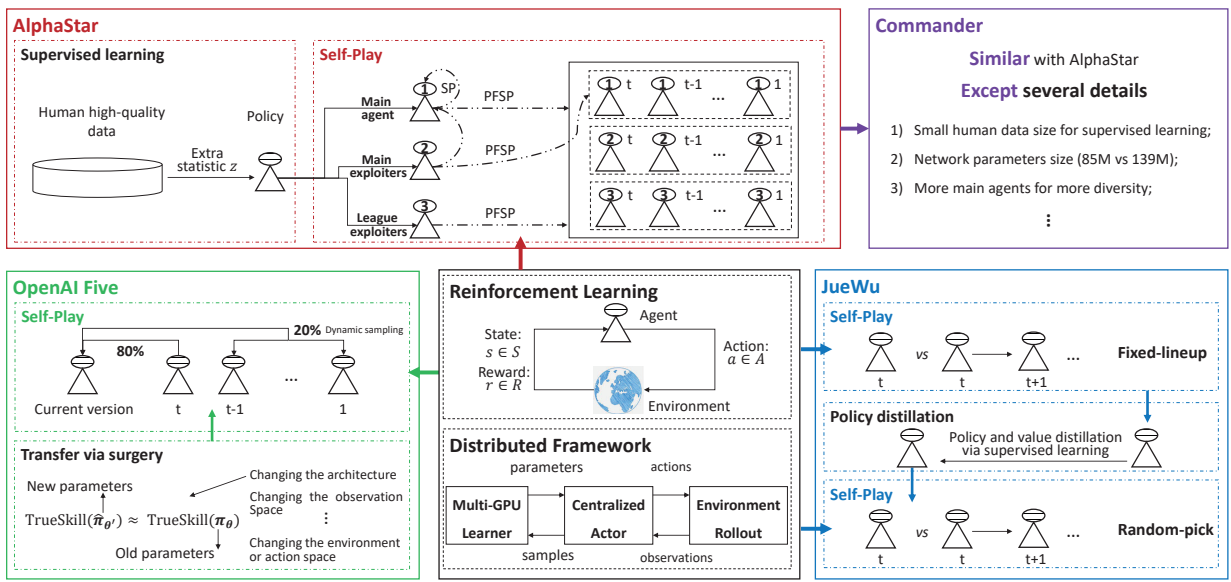}
   \end{center}
   \caption{A brief AI framework for typical RTS games.}
   \label{AlphaStar}
\end{figure*}

\subsection{Population based Training}
Population based training maintains a population of agents, which consists of two important components to direct learning process: sample teammates and opponents for an agent to generate data, reset and perturb hyper-parameters and transformation parameters for underperforming agents based on training process.

When collecting training data for agent policy optimization, a sampling method based on Elo of agents is utilized.
It encourages agents with similar skill (Elo scores) to be teammates and opponents, ensuring that the outcome of a game is sufficiently uncertain so as to guild
agent learning useful policy.
Since conventional Elo calculation method is designed without considering agent cooperation, FTW makes an assumption that rating of a team can be decomposed as sum of skills for a team.
With above assumption, Elo for each agent can be obtained following regular Elo optimization approach.

After training a generation of population, hyper-parameters like learning rate and transformation parameters between the intermediate rewards and global rewards for underperforming agents are reset and perturbed by using the better performing agents as reference.
More specifically, if an agent with a team cannot win another agent with a team (e.g., 70\% wining rate), the losing agent copies the policy, reward transformation, and hyper-parameters of the better agent, and then probabilistically perturb the inherited values with a small range, e.g., $\pm 20\%$ with a probability of 5\%.
The above exploration process helps to find better hyper-parameters and transformation parameters.

\section{RTS Game AIs}\label{sec6}
RTS game, as a typical kind of video game, owns tens of thousands of people to fight against each other, which naturally becomes a testbed for human-computer gaming.
Furthermore, RTS games are usually complex environment, which capture more nature of real world than previous games, making breakthrough of such games more applicable.
AlphaStar, developed by DeepMind, uses general learning algorithms and reaches grandmaster level for all three races for StarCraft, which also outperforms 99.8\%
human players who are active on the European server (about 90000 players).
Commander, as a lightweight computation version, follows the same training architecture of AlphaStar, which uses order of magnitude less computation and beats two grandmaster players in a live event.
OpenAI Five aims to solve Dota2 game, which is the first AI system that defeats the world champions at an esports game.
As a relatively similar esports game with Dota2, Honor of Kings shares most similar challenges, and JueWu becomes the first AI system that can play full RTS games instead of restricting the hero pool.
A brief introduction is shown in figure \ref{AlphaStar}.

\subsection{Basic Techniques for RTS Game AIs}
To handle complex RTS games, reinforcement learning accelerated by distributed framework becomes a basic tool.
Different from the distributed frameworks designed for Suphx and DouZero, a larger data throughput framework is designed because a huge interaction with environment is required.
Previous distributed reinforcement learning mainly maintains two important modular: parallel environments with each embedded an actor to generate actions, and learners to consume data collected by the environments for policy updating.
With such a distributed framework, plenty of time is wasted because in each environment, a model inference should be conducted for a single action.
Current distributed reinforcement learning performs centralized model inference for states collected from multiple environments and distributes actions for each environment as shown in the bottom of figure \ref{AlphaStar}.
Based on the learner-centralized actor-environment architecture, model inference time is largely reduced, which will save time for big models used for complex games.

\subsection{Training for AlphaStar}
Training of AlphaStar consists of two main steps: supervised learning to initialize agent parameters and multi-agent reinforcement learning to improve the agent.
In supervised learning, a high quality dataset is collected to train the agent parameters.
The dataset consists of 971000 replays from human players, whose MMR scores are greater than 3500, i.e., in the top of 22\% of players.
Since there are three races for StarCraft, AlphaStar trains one agent for each race.
To fully explore human experience especially in the game beginning where little combat feedback can be obtained, AlphaStar extracts a statistic variable to condition the policy, and adopts KL divergence between human actions and the policy¡¯s outputs to assist learning.
Such statistic variable encodes each player's first 20 buildings and units, which reflects a type of opening strategy for AlphaStar.
After above supervised training, AlphaStar fine-tunes the policy using a subset but more professional human player data (with MMR above 6200), which improves the policy by 9\% percentage when fighting against built-in elite bot.

After supervised learning for agent initialization, a multi-agent reinforcement learning framework with league training is developed, so as to alleviate the game-theoretic challenges such as cycles between strategies.
We firstly introduce agent types in the league, and then elaborate how to train different agents.
The league has three types of agents for each race: main agent, main exploiter and league exploiter.
Training of those agents lies in how to select opponents in the league and whether or not to reset the learned parameters.
Specifically, opponents of main agent are main agent itself and all agents in the league, so as to be strong enough for final testing.
Opponents of main exploiter are current main agent and previous main agent versions, to find weaknesses of the main agent.
Opponents of league exploiter are all agents in the league, to discover possible weaknesses of the entire league.
With main exploiter and league exploiter added in the league, training of main agent can properly overcome the weakness of itself and in the league.

When deciding sampling probabilities of opponents for different type of agents, an improved version of fictitious self-play called prioritized fictitious self-play is designed, which selects opponents based on wining rate against the agent, instead of a uniform mixture of opponents.
Detailed probability distribution and calculation can be find in original paper.
Noted that when a generation of main exploiter or league exploiter agent is obtained, it is periodically reinitialized to supervised learned agent, so as to extend diversity of the league.

\subsection{Training for OpenAI Five}
Training of OpenAI Five is based on distributed self-play deep reinforcement learning.
With their distributed learning system, OpenAI Five successfully extends the learning batch size to be 2,949,120 time steps, which are important for training.
When performing self-play to generate training data, agent plays against itself for 80\% of the games and against past versions for 20\% of the games.
Modifying conventional self-play in above way avoids strategy collapse and ensures the learned agent being robust to a wide range of opponents.
To effectively sample opponents from a large number of past versions, OpenAI Five maintains a score for each agent and changes the score based on the wining signal of training trajectories.
This strategy makes sure a dynamic sampling is performed to select useful agents to play against.

Another key factor for success of OpenAI Five is a tool called continual transfer via surgery, which adjusts parameters of a learned model for adapting to new version of Dota2.
Such tool is essential because Valve company usually publishes a new version of Dota2 every a few months, resulting performance degradation of the learned model.
Even though a new model can be trained from scratch,  the time is limited and the resource consumption is intolerable.
What's more, the designed tool makes training of the agent more efficient because model parameters and architectures can be adjusted based on performance in training process.
Parameters transfer obeys a basic rule, i.e., TrueSkill of new agent (new parameter space) matches that of already learned agent.
Based on such principle, OpenAI Five develops different methods for changes of the architecture, observation space, action space and so on.

\subsection{Training for JueWu}
Training of JueWu is similar with that of OpenAI Five, where no human player data is utilized for agent initialization.
However, to play with a hero pool of full RTS game instead of restricting the selection of heros, JueWu develops new training framework compared with the basic form of OpenAI Five.
More specifically, training of JueWu consists of three main steps: fixed-lineup training, multi-teacher policy distillation and random-pick training, followed by a MCTS based approach for learning to draft.

Considering self-play of massive disordered agent combinations makes training of an agent a very hard task, JueWu adopts a curriculum based training scheme: firstly using fixed-lineup and then utilize random pick.
Several fixed lineups without hero repeat are carefully selected, based on which, distributed reinforcement learning is performed to train several teacher agents.
To generate such lineups, JueWu analyses vast amount of human player data, and selects relatively balanced teams.
Based on the teacher agents, a policy distillation is conducted to learn a bigger student agent.
The distillation is modeled as a supervised learning framework to minimize the difference between outputs of teacher and student models, i.e., Shannon¡¯s cross entropy between action distributions and Euclidean distance between value estimations.
Finally, based on the student agent, another distributed reinforcement learning is applied for random pickups.
Student agent, learned from fixed-lineup and served as initialization of random pick, largely reduce training difficulty.

A very important and interesting part in RTS games like Dota2 and Honor of Kings is hero drafting to form two teams.
JueWu proposes a MCTS and neural network based approach to handle the problem of huge combination of agents, i.e., more than $10^{11}$.
The motivation of using neural network in MCTS is similar with AlphaGo Zero, namely estimates the value of the expanded node more accurate and to avoid a complete rollout, which is very time consuming.
Unlike OpenAI Five, the terminal state of draft is not the end of a game, so wining or losing signal cannot be obtained.
To construct the dataset to training value estimation network, the label, i.e., wining signal, should be obtained.
To solve this problem, JueWu collects another dataset, which performs plenty of matches using randomly selected teams with the learned reinforcement learning model.
Then, a lineup-wining result dataset is developed, based on which, a wining prediction network can be trained and used as signal for value network training labels.

\subsection{Training for Commander}
Similar with AlphaStar, Commander adopts a very similar training framework for StarCraft agent learning, i.e, supervised learning followed by multi-agent reinforcement learning.
The main differences are several important details, which makes Commander beats two professional players with order of magnitude less computation.
Firstly, Commander uses a much smaller human player dataset, based on which, learning rate, batch size, multi-stage training and network structure are carefully designed for supervised learning.
In multi-agent reinforcement learning, Commander devises the training loss, and uses more main agents for more diversity, which improves the learning efficiency.

\subsection{Training Difference}
Nowadays, deep reinforcement learning accelerated by distributed learning becomes a general method to train high performance AIs.
Apart from this, the four typical AIs, i.e., AlphaStar, OpenAI Five, JueWu and Commander share several differences.

Firstly, to train each generation of agents, those AIs utilize different self-play or revised self-play mechanisms.
In JueWu and OpenAI Five, relatively simple self-play is performed to train each generation of agents.
To avoids strategy collapse and ensure the learned agent being robust to a wide range of opponents, usually a certain percentages of past versions are selected as opponents.
This selection can be specially designed instead of using fictitious self-play, i.e, uniformly select past versions.
For example, OpenAI Five selects past versions with 20\% of rollout games.
AlphaStar utilizes a prioritized fictitious self-play mechanism to select opponents, based on which, relatively hard agents and agents with similar levels are more likely to be chosen.
What's more, AlphaStar and Commander adopt league training, which is a power mechanisms to enhance self-play for more diverse agents learning.

Secondly, purely based on reinforcement learning usually requires a huge computational resources because of its trial and error mechanism, so those AIs utilize human player data to assist reinforcement learning.
In AlphaStar and Commander, supervised learning based on high quality data is performed to initialize policy networks, so as to provide good and diverse initialization for reinforcement learning.
What's more, statistics are extracted from human data to constrain the policy in reinforcement learning stage, which helps a lot based on the ablation study in their papers.
In JueWu, human data are not used for policy initialization.
Instead, the data is used to analyze the hero lineups, so as to provide relatively balanced teams for first learning stage, i.e, self-play reinforcement learning with fixed-lineup.
In OpenAI Five, no human data are utilized, and OpenAI just utilize self-play reinforcement learning for policy training, using huge computational resources for over 10-month training.

Thirdly, several new techniques are developed to overcome some challenge problems in the games.
Different from population based training in FTW, AlphaStar maintains a league for agent training, where different types of agents are responsible for different tasks.
Even though being heuristic, league based multi-agent training provides a very useful idea for complex realtime games with game-theoretic challenges.
Continual transfer via surgery, as an effective tool to make full use of currently learned model for changing environment, is very useful because real world environment is inevitably changing through time.
Such a technique can largely reduce computation cost, and change models when it is necessary.

\section{Techniques comparison}\label{sec7}
Based on current breakthrough of human-computer gaming, techniques can be roughly divided into two categories: tree search assisted by deep neural network, and advanced self-play with distributed deep reinforcement learning.

\subsection{How to Reach Nash Equilibrium?}
Nash equilibrium \cite{Nash}, an important concept in game theory, is the best strategy for any players no matter what strategies the other players chose.
Due to the above characteristic, researchers have paid much attention on approaching Nash equilibrium \cite{Nash1,Nash2}.

Tree search methods have long been a mainstream for turn based games.
Typical methods such as min-max search, MCTS and CFR are classical algorithms that can approach Nash equilibrium, so those techniques are widely utilized in games such as chess and limit poker.
However, when facing complex environments such as Go and HUNL, the calculation of Nash equilibrium is untraceable because of the huge game tree complexity.
To handle such problem, properly restricting depth and width of the game tree becomes a very important strategy, where deep learning can be used.
For example, AlphaGo series train policy and value networks so as to pay more attention on valuable nodes to be expanded and to evaluate nodes expanded, respectively.

In complex real time video games, we cannot easily draw lessons from tree search methods because of challenges such as long time horizon and complex action space.
Fictitious self-play \cite{FSP} provides an evolutionary strategy for agent learning, which can approach the Nash equilibrium in certain types of games.
However, computation of fictitious self-play for complex game is high, so researchers develop various self-play strategies, and uses distributed reinforcement learning to learn each generation of agents.
Even though theoretical guarantee for Nash equilibrium is absent, professional level AIs can be trained by properly overcoming game-theoretic challenges.
For examples, OpenAI Five play against itself for 80\% of the games and against past generations for 20\% of the games by their winning rate against current version.
AlphaStar designs three types of agents to enhance self-play, where each type of agent performs confrontation with certain opponentes, so as to gradually improve performance of the main agents without desperation or just learn a narrow of policies.

\subsection{How to Become General Technology?}
\begin{figure}
   \begin{center}
   \includegraphics[width=0.475\textwidth]{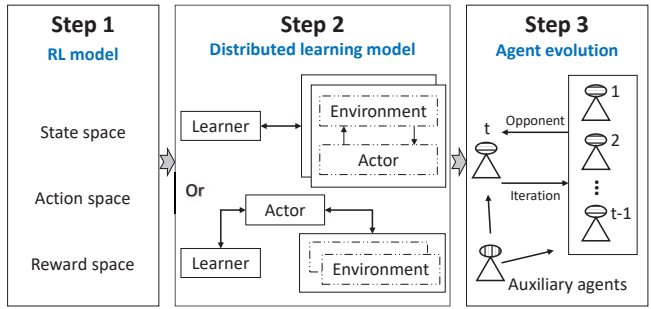}
   \end{center}
   \caption{Steps for a general technology to train AIs.}
   \label{general}
\end{figure}

Considering real world games are mostly real time with a lot of decisions to be made, and players usually form their decisions not in an iterative manner, tree search based methods are not so easy to be implemented in very complex games.
However, advanced self-play with distributed learning can be a general solution due its simple implementation and performance guarantee such as success of AlphaStar and OpenAI Five.
Generally, there are three steps to train an AI based on this technique, as shown in figure \ref{general}.

Firstly, the task can be properly modeled as a reinforcement learning framework, which consists of
several key factors.
Usually, determining the state space and action space is the most important part.
The former provides information for neural network input, which should be rich enough for a suitable decision and lean enough to reduce computation burden.
The latter reflects how to drive environment transfer.
Too complex of action representation will increase learning difficulty, but too simple design will make the agent unable to reach professional level due to action limitations.
What's more, When performing reinforcement learning, how to design reward space is another key factor because it is the task signal to learn each generation of agents.
Too sparse reward under long time horizon game will greatly increase learning difficulty, but designing immediate rewards to guild agent pursuing task reward needs a lot of human experience.

With above factors, one can design or adopt reinforcement learning algorithms such as Q learning, Advantage Actor Critic \cite{A3C}, Proximal Policy Optimization \cite{PPO} for agent learning.
Usually, deep neural networks are specially designed to transform input state information to output action, e.g., auto-regressive policy to deal with structured and combinatorial action space in AlphaStar.
To accelerate reinforcement learning, distributed learning should be carefully designed, based on the model inference cost to drive rollout, communication cost among machines to transfer data, and most importantly the machine configure such as GPU and CPU ability.
For example, when the model is relatively small and the inference cost is low but there are no fast connections for data transformation between and within machines, one can chose distributed framework like in FTW.
Nowadays, Tensorflow\footnote{https://www.tensorflow.org/}, Pytorch\footnote{https://pytorch.org/} and several tools such as Ray \cite{Ray} and Horovod \cite{Horovod} can easily achieve multiple machines distributed learning with minimal code changes compared with that in single machine \cite{DisComparison}.

Finally, since each generation of agent can be trained based on distributed reinforcement learning, a last step is to design self-play based mechanism for agent evolution.
Currently, plenty of heuristic approaches have been developed.
For example, AlphaStar uses three types of agents with each type selects different opponents, based on which, all types of agents evolve to make the main agent stronger.
Overall, previous evolution strategies for self-play are mostly heuristic, and one can design strategies based on the game faced, so as to improve the agent ability.

\section{Challenges and Future Trends}\label{sec8}

Even though big progress has been made in human-computer gaming, current techniques still suffer from challenges like relying much on computational resources, which will inspire future researches.

\subsection{Big Model}
Nowadays, big model, especially pretrained big model, is emerging from natural language processing to computer version, from single modality to multiple modalities \cite{BigModel,M6}.
Those models have proved great potential for downstream tasks even in zero-shot settings, which is a big step for exploring artificial general intelligence.
For example, OpenAI developed Generative Pre-trained Transformer 3 (GPT-3) \cite{GPT}, which has more than 175 billion parameters and displays promising performance in various language related tasks.
However, big model in games is largely absent, and current models for complex games are much smaller than those big models.
As shown in Table \ref{parameters}, AlphaStar and OpenAI Five only have 139 million and 159 million parameters, respectively.

\begin{table}[htbp]
\begin{center}
\caption{Parameter sizes of current AIs and pretrained models.}
\label{parameters}
\begin{tabular}{l l}
\hline
Models     &   Parameter size   \\
\hline
JueWu \cite{Honor}   &   17 million   \\
Commander \cite{Commander}   &   49 million    \\
AlphaStar \cite{AlphaStar}  &   139 million   \\
OpenAI Five \cite{OpenAIFive} &   159 million    \\
\hline
GPT-3 \cite{GPT}  &   175 billion     \\
Megatron-Turing NLG\tablefootnote {https://www.microsoft.com/en-us/research/blog/using-deepspeed-and-megatron-to-train-megatron-turing-nlg-530b-the-worlds-largest-and-most-powerful-generative-language-model/} &   530 billion \\
\hline
M6-10T \cite{M610T} &   10 trillion \\
\hline
\end{tabular}
\end{center}
\end{table}

Considering big model is a relatively good exploration for artificial general intelligence, how to design and train big model for AI in human-computer gaming, may provide a solution for those sequential decision making problems.
To give such an attempt, we think at least two problems should be carefully considered.

Firstly, unlike in natural language processing problems, tasks for games are very different, so how to make clear of training goal is key step for big model.
For example, in StarCraft, players need to build force with at most 200 units to fight against enemies, but in Dota2, five heros are working together to defeat another five heros.
Even through distinct actions or skills are required for different games, the mechanism of playing a game is similar, i.e., extract useful information of image streams and make a decision based on current situation.
So a possible breakthrough point is to learn high-level strategic situation, so as to provide information for decisions.
Noted that other goals for training big model are welcomed as long as they can provide general and useful information for making decision.

Secondly, since some games are hard and some games are easy, how to design a suitable training mechanism is difficult.
It should handle various kinds of games and make sure the learning do not degenerate, e.g., not forgetting the representation ability.
Continual learning provides a tool for such problem \cite{CLNLP,CLCV}, but there are still several issues need to be carefully handled.
Since training a high level game AI is an evolution process which needs self-play or other iterative learning, how to properly embed evolution in to above learning mechanism is a problem that has never been faced.
On the other hand, different games share similar characteristics to some extent, how to establish connection between them when performing training is a key factor to reduce complexity and meanwhile promote performance.

\subsection{Low Resources AI}
To train professional level AIs for complex environments, usually a large computational resources are required.
As shown in Table \ref{resources}, we can find a huge resources devotion to train an AI.

\begin{table}[htbp]
\begin{center}
\caption{Computational resources for professional AIs.}
\label{resources}
\begin{tabular}{l l}
\hline
AIs     &   Resources   \\
\hline
AlphaZero   &   5000 v1 TPUs and 16 v2 TPUs for 13 days   \\
Libratus    &   25 million core hours    \\
OpenAI Five &   $770\pm50$PFlops/s$\cdot$day for 10 months    \\
AlphaStar   &   192 v3 + 12 128 core TPUs, 1800 CPUs for 44 days   \\
\hline
\end{tabular}
\end{center}
\end{table}

One question naturally raises that if it is possible to train a professional level AI with limited resources.
One intuitive idea is to bring in more human knowledge to assist learning \cite{Prior1}.
For example, incorporating prior knowledge as constraints or loss functions for conventional machine learning algorithms.
Since current breakthroughs on games are mostly relying on reinforcement learning which is low sample efficient, how to achieve sample efficient reinforcement learning based on human knowledge is a future direction \cite{PriorRL1,PriorRL2}.

On the other hand, training a professional agent is usually an evolution process, which iteratively learns hundreds of models.
For example, In AlphaStar, almost 900 different players are created, with each one maintains a specific kind of task.
So how to reduce such iteration seems to be an effective medium for reducing computational resources.
Current approaches, mainly based on self-play, are mostly heuristic by selecting suitable opponents for current generation of agent.
If theoretical and easy to calculate evolution strategies are developed, it will be a key step for low resources AIs.

\subsection{AI Evaluation}
Most games in real world are in-transitive, i.e., transitive and in-transitive parts are co-existing \cite{Transitive}.
The in-transitive characteristic makes precise evaluation of agent a difficult problem.
Current human-computer gaming usually utilizes winning rate (against professional human players) based evaluation criteria, as shown in Table \ref{evaluation}.
However, such evaluation is relatively rough especially under limited tests for in-transitivity games.
\begin{table}[htbp]
\begin{center}
\caption{Evaluation of typical AIs.}
\label{evaluation}
\begin{tabular}{l l}
\hline
AIs     &   Resources   \\
\hline
AlphaGo Zero  &  previous AlphaGo series\tablefootnote{Including AlphaGo Master, a previous version of AlphaGo Zero that defeated strongest human professional players by 60$-$0 in online games.}   \\
Suphx   & 99.99\% of all the officially ranked human players \\
Libratus    &   Four top human specialist professionals    \\
OpenAI Five &   Professional teams with world champions OG  \\
AlphaStar   &   99.8\% of ranked human players     \\
JueWu   & 95.2\% win rate against professional players \\
\hline
\end{tabular}
\end{center}
\end{table}

Theoretically, Nash equilibrium is a relative conservative solution due to not considering weakness of opponents \cite{opponent1,opponent2,opponent3}.
Still, it is a best solution for any kinds of opponents in non-cooperative games.
Accordingly, how to evaluate the distance between obtained solution with Nash equilibrium solution is an important problem.
It may helps us figure out if AlphaZero reaches the Nash equilibrium and can not be beaten by any humans.

On the other hand, current ranking methods for human and AIs are based on their battle records with mainly generative calculation based methods such as Elo \cite{evaluation1,evaluation2,evaluation3}.
However, under in-transitive games, such calculation is inexact.
Moreover, win rate is just one of the evaluation metrics, and it may be not enough to reflect all the aspects of an agent.
Accordingly, how to develop a systematic evaluation criteria for most games can be an important and open problem.

\subsection{New Challenging Games}
After the breakthrough of AlphaStar, researchers are looking for new games for advancing decision making intelligence, e.g. football.
In our opinion, current games with big progress are mostly symmetrical in ability.
Even through games like StarCraft and Dota2 look like asymmetric because there are three distinct races with different forces in StarCraft and plenty of heros with diverse skills in Dota2, those games share a common characteristic of balance for different choices.
This is important for games being popular for humans, e.g., being an esport.

On the contrary, real world is full of asymmetric games, and it is almost unable to find a strictly symmetrical game in our surroundings \cite{AsyGames}.
So a practical issue raises, it maybe a good direction to design asymmetrical games (mostly asymmetrical in ability), so as to develop decision making intelligence for real world problems.
However, there are few environments of asymmetric games, and researchers are paying much little attention on developing techniques for those kinds of testbeds \cite{LanQiu}.
We argue that previous training frameworks, especially self-play with distributed learning, can not deal with such senecios, because a two player asymmetric game has very different strategies for different sides, and self-play based mechanisms may not work well.

\begin{figure}
   \begin{center}
   \includegraphics[width=0.4\textwidth]{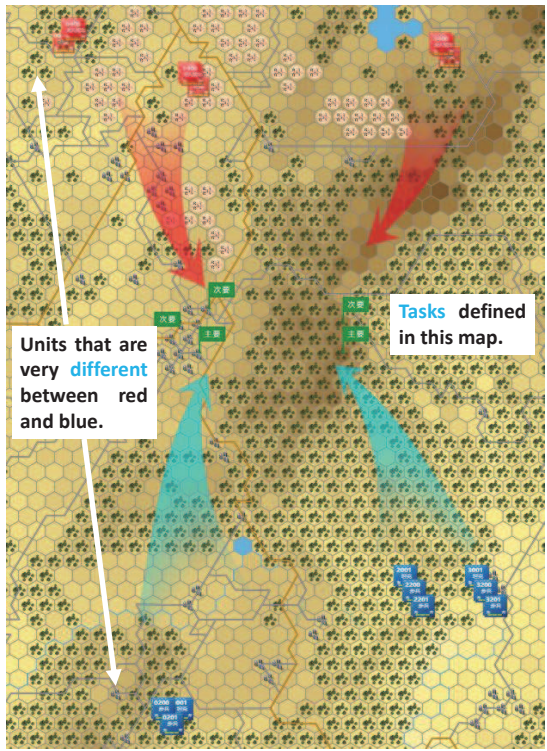}
   \end{center}
   \caption{A screenshot of wargame.}
   \label{wargame}
\end{figure}

Wargame, is a popular confrontation game, as shown in figure\footnote{Come from http://wargame.ia.ac.cn/main} \ref{wargame}, where two players (red and blue) with each controls a collection of combat units fight against each other \cite{Yin}.
Based on several settings of Wargame, two sides are asymmetrical in ability and usually the power of red one is weaker than that of blue one.
Considering Wargame is a complex game like AlphaStar that faces imperfect information, long time horizon, in-transitive game and multi-agent cooperation, and its distinctive asymmetric game characteristic, it may be a new testbed for AI in human-computer gaming.

\section{Conclusion}\label{sec9}
In this paper, we have summarized and compared techniques of current breakthroughs of AIs in human-computer gaming.
By comparing the techniques utilized, we illustrate the mainstream frameworks and techniques for developing professional level AIs.
More importantly, we try to raise challenges of current decision making techniques, hoping to inspire future directions in the field.
Through this brief survey, we hope beginners can quickly familiar with techniques, challenges and opportunities in this exciting field, and researchers on the way can be inspired for deeper study.

\bibliographystyle{IEEEtran}
\bibliography{mybibfile}

\end{document}